\documentclass[10pt,twocolumn,letterpaper]{article}

\usepackage{cvpr}              
\definecolor{cvprblue}{rgb}{0.21,0.49,0.74}
\usepackage[pagebackref,breaklinks,colorlinks,allcolors=cvprblue]{hyperref}
\usepackage{times}
\usepackage{latexsym}
\usepackage{enumitem}
\usepackage{array}

\newcommand*{\affaddr}[1]{#1} 
\newcommand*{\affmark}[1][*]{\textsuperscript{#1}}

\usepackage{booktabs}
\usepackage{makecell}
\usepackage{graphicx}
\usepackage[T1]{fontenc}

\usepackage[utf8]{inputenc}

\usepackage{microtype}

\usepackage{inconsolata}

\usepackage{graphicx}
\usepackage{algorithm}
\usepackage{algorithmic}
\usepackage{amsmath}
\usepackage{amssymb}
\usepackage{booktabs}
\usepackage{caption,subcaption}
\usepackage{makecell}
\usepackage{tcolorbox}
\usepackage{graphicx}
\usepackage{tabularx}
\usepackage{array}
\usepackage{afterpage}
\usepackage{arydshln}
\usepackage{array} 
\newcolumntype{C}[1]{>{\centering\arraybackslash}m{#1}}
\usepackage{newfloat}
\usepackage{listings}

\usepackage{siunitx}
\usepackage{booktabs}
\usepackage{multirow}

\usepackage{siunitx}

\def\paperID{18} 
\def\confName{CVPR}
\def\confYear{2026}

\title{Seeing Straight: Document Orientation Detection for Efficient OCR}

\author{Suranjan Goswami\affmark[1], Abhinav Ravi\affmark[2], Raja Kolla\affmark[2],  Ali Faraz\affmark[2], Shaharukh Khan\affmark[2] \\ 
\textbf{Akash\affmark[1], Chandra Khatri\affmark[2] and Shubham Agarwal\affmark[2]}  
\\ 
\\
\affaddr{\affmark[1]OLA Electric, Bangalore, India}\\
\affaddr{\affmark[2]Krutrim AI, Bangalore, India}\\
\texttt{\{suranjan.goswami, akash.shyam\}@olaelectric.com} \\
\texttt{\{raja.kolla, ali.faraz, shubham.agarwal1\}@olakrutrim.com} 
}

\begin{document}
\maketitle
\begin{abstract}
Despite significant advances in document understanding, determining the correct orientation of scanned or photographed documents remains a critical pre-processing step in the real world settings. Accurate rotation correction is essential for enhancing the performance of downstream tasks such as Optical Character Recognition (OCR) where misalignment commonly arises due to user errors, particularly incorrect base orientations of the camera during capture. In this study, we first introduce OCR-Rotation-Bench (ORB), a new benchmark consisting of 1863 images for evaluating OCR robustness to rotations, comprising (i) ORB-En, built from rotation-transformed structured and free-form English OCR datasets, and (ii) ORB-Indic, a novel multilingual set spanning 11 Indic mid to low-resource languages. We also present a fast, robust and lightweight rotation classification pipeline built on the vision encoder of Phi-3.5-Vision model with dynamic image cropping, fine-tuned specifically for 12-class rotation task in a standalone fashion. Our method achieves near-perfect 98\% and 96\% accuracy on identifying the rotations respectively on both the datasets. Beyond classification, we demonstrate the critical role of our module in boosting OCR performance: closed-source (up to 20\%) and open-weights models (up to 4x) in the simulated real-world settings.  
\end{abstract}    

\begin{figure*}[htbp]
\centering
\includegraphics[width=1\textwidth, keepaspectratio]{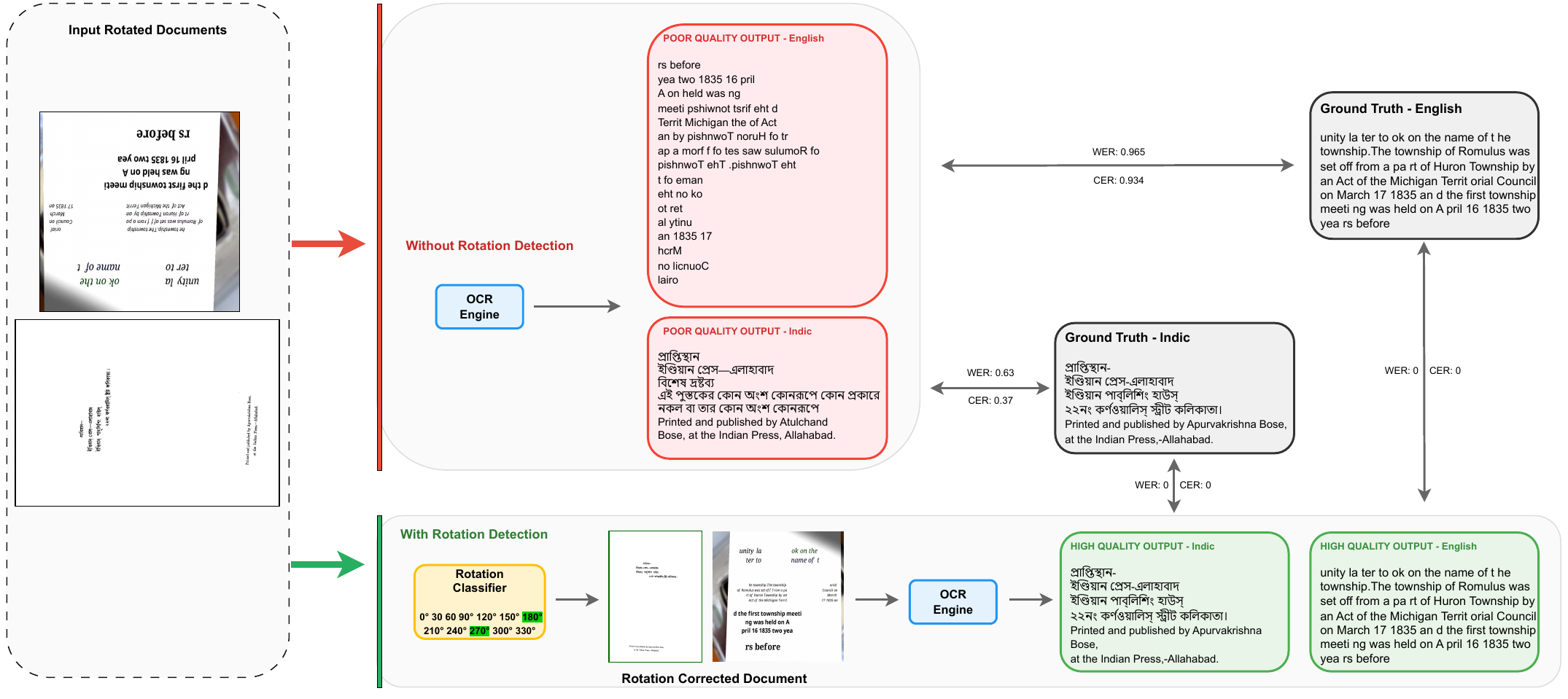}
\caption{\textbf{OCR workflow comparison} with and without rotation correction on sample English and Indic documents. Errors due to misalignment such as repetitions and hallucinations (in red) are mitigated by our module, yielding accurate outputs (in green).}
\label{fig1}
\end{figure*}

\section{Introduction}
\label{sec:intro}

Recent advancements in Vision Language Models (VLMs)~\cite{bai2023qwenvl,chen2024far,lu2024deepseek-vl,beyer2024paligemma,abdin2024phi2,xue2024xgen-mm,xiao2024florence} have demonstrated
impressive capabilities in general purpose visual-linguistic understanding. Models such as DocOwl series~\citep{hu2024mplug, docowl2_2024}, DocLLM~\citep{wang2023docllm} and Donut~\cite{kim2022donut} have significantly advanced document understanding tasks. 


One of the key applications of such systems is Optical Character Recognition (OCR), which transforms scanned or photographed document images into machine-readable text. However, OCR pipelines often rely on clean inputs without dedicated mechanisms to ensure robust orientation handling. Figure~\ref{fig1} illustrates the impact of rotation misalignment on OCR performance for both an English as well as a Bengali script document. In the absence of rotation correction, the model exhibits recognition errors such as repetitions and hallucinations. The same OCR model produces accurate and coherent text, effectively mitigating these issues when provided with upright images.  

Even the benchmark evaluations are typically conducted on curated datasets like RVL-CDIP~\cite{harley2015icdar}, FUNSD~\cite{jaume2019funsd}, and 
DocVQA~\cite{mathew2021docvqa}, \textit{inter alia}, where documents are upright and well aligned. Orientation is usually assumed or corrected as a pre-processing step, limiting their robustness in practical scenarios. In contrast, real-world documents such as receipts, forms, handwritten notes, and IDs are often captured via mobile phones or scanners under uncontrolled conditions, resulting in unintended rotations. These misalignments can significantly degrade OCR and information extraction accuracy. Furthermore, existing evaluation datasets are heavily skewed toward English or Latin-script documents, with limited insights for multilingual settings, especially for Indian languages. Resources like Wikisource\footnote{\url{https://wikisource.org/}}, an open-access repository of digitized literary works, present a promising avenue for constructing large-scale multilingual OCR benchmarks. However, the lack of standardized, rotation-diverse, and multilingual datasets remains a bottleneck in evaluating and improving  generalization of current OCR engines. 

To address these gaps, we propose a lightweight rotation classification module that operates as a pre-processing step to OCR in noisy, real-world settings. Specifically, we fine-tune the vision encoder of Phi-3.5-Vision-Instruct~\cite{abdin2024phi3} as a standalone module for this task, improving OCR robustness. We formulate orientation correction as a twelve-class classification task over 30 degree angle variations, extending prior works ~\citep{unnikrishnan2009combined}, effective for most documents which are typically scanned or captured in vertical or horizontal orientations. Our method is efficient, accurate, and generalizes across languages and scripts, enabling reliable alignment of rotated document images before text recognition. 

In addition, we further introduce a novel multilingual OCR benchmark spanning 11 mid- to low-resource Indic languages, constructed from high-quality document-text pairs sourced from Wikisource. To complement this, we incorporate three standard English OCR datasets, transformed with controlled rotations to simulate real-world misalignment scenarios. Together, these form a comprehensive benchmark for evaluating OCR robustness to rotation. We evaluate our rotation correction module alongside state-of-the-art OCR models and demonstrate consistent improvements in performance across both English and Indic scripts. Our key contributions could be summarized as:
\begin{itemize}[noitemsep] 
\item We present OCR-Rotation-Bench (ORB), a novel benchmark for evaluating OCR robustness to practical image rotation scenarios. ORB consists of two subsets: (i) ORB-En, derived from rotated versions of 3 standard English OCR datasets of total 897 images, and (ii) ORB-Indic, a curated set of 966 rotated real-world multilingual document images with OCR ground truth (GT) annotation across 11 Indic languages.
\item We propose a real-time, lightweight rotation classification method based on a standard vision encoder, deployable across diverse document pipelines as a pre-processing step.
\item We also conduct an exhaustive evaluation of downstream OCR performance, demonstrating significant gains in accuracy when a correction module is integrated. We will publicly release all the related artifacts
to support reproducibility and community benchmarking.

\end{itemize}

\begin{table*}[htbp]
\centering
\resizebox{\textwidth}{!}{
\begin{tabular}{ccccccccccc}
\hline
\textbf{Hindi} & \textbf{Bengali} & \textbf{Tamil} & \textbf{Malayalam} & \textbf{Telugu} & \textbf{Marathi} & \textbf{Kannada} & \textbf{Gujarati} & \textbf{Punjabi} & \textbf{Oriya} & \textbf{Assamese} \\
\hline
86 & 90 & 89 & 89 & 89 & 86 & 89 & 90 & 89 & 86 & 83 \\
\hline
\end{tabular}
}
\caption{Distribution of Indian languages represented in the ORB-Indic benchmark.}
\label{tab:language-distribution}
\end{table*}
\section{Related Work}
\label{sec:related}
\textbf{Rotation Correction in Document Pre-processing:} Orientation correction is a fundamental step in OCR pipelines, particularly for documents captured via mobile devices or scanners. Classical methods such as projection profiles, Hough transforms, and connected component analysis estimate dominant text skew using low-level visual features ~\cite{bao2022adaptive}. While effective on clean, structured inputs, these approaches degrade under noise, script diversity, or low resolution. ~\citet{unnikrishnan2009combined} formalized rotation into four canonical classes: $0^\circ$, $90^\circ$, $180^\circ$, and $270^\circ$. ~\citet{hu2001automatic} explored symmetry-based textual analysis in natural scenes, illustrating the effectiveness of spatial patterns for orientation detection. While recent deep learning approaches explore implicit orientation correction ~\cite{zhao2022transformer}, they remain limited to English. In contrast, our work isolates and enhances the rotation step, enabling language-agnostic benchmarking and seamless integration into existing OCR workflows.

\textbf{Optical Character Recognition (OCR):} OCR systems have progressed from traditional rule-based engines like Tesseract~\cite{smith2007overview} to deep learning models that integrate visual and sequential components. Modern approaches such as docTR~\cite{liao2023doctr}, and TrOCR~\cite{li2021trocr} demonstrate strong performance on both printed and handwritten text in English, leveraging transformer-based visual encoders and pretrained language decoders. However, these models typically assume upright inputs, and even minor rotations can significantly degrade recognition accuracy, highlighting the need for explicit orientation correction. On the other hand, recent multilingual OCR systems, such as the closed-source Chitrapathak, GPT-4o,  Gemini-2.5 Flash and Gemini-2.5 Pro\footnote{
\begin{minipage}[t]{\linewidth}
\url{https://ai-labs.olakrutrim.com/} \\
\url{https://platform.openai.com/docs/models/gpt-4o/} \\
\url{https://deepmind.google/models/gemini/}
\end{minipage}
} have broadened language support. 
Recently released open-source solution like Nanonets~\cite{nanonets2024ocr} claims to provide industrial-grade OCR pipelines,  
utilizing fine-tuned Qwen-VL backbones~\cite{qwen2023vl}, effectively bridging foundational vision-language models with document-specific applications in the multi-lingual setting. We show the performance of these models on both structured and free form generations with our trained module.

\textbf{VLMs for Document Understanding:}
Recent VLMs such as the LLaVA family~\citep{liu2024visual, liu2024improved} have advanced multimodal alignment by coupling vision encoders~\citep{radford2021learning, zhai2023sigmoid, tschannen2025siglip} with large language models via \textit{visual instruction tuning}. Several successors~\citep{lu2024deepseek, laurenccon2024matters, tong2024cambrian, xue2024xgen, team2024gemma} follow similar paradigms, excelling in translation, captioning, and visual reasoning. Document-centric variants such as DocOwl~\citep{hu2024mplug, docowl2_2024}, DocLLM~\citep{wang2023docllm}, and Donut~\citep{kim2022donut} with emerging datasets like BigDocs~\citep{bigdocs2024}, 
show promise but remain English-centric and rotation-agnostic.

Support for Indic languages in VLMs remains limited. Models such as Chitrarth~\citep{khan2025chitrarth} and Chitranuvad~\citep{khan2025chitranuvad}, built on Krutrim~\citep{kallappa2025krutrim}, represent early efforts to support 10 Indian languages and low-resource translation respectively. Others, including PALO~\citep{maaz2024palo}, Maya~\citep{alam2025behind}, and Pangea~\citep{yue2024pangea}, partially supports select languages like Hindi, Bengali, Tamil, and Telugu but remain general-purpose VLMs. However, none explicitly address document understanding or orientation robustness in low-resource Indian contexts, which our work directly addresses. 
\section{Model Architecture}
\label{sec:arch}

We define the architecture of our rotation classifier model as the following. 
Let \( x \in \mathbb{R}^{H \times W \times 3} \) denote an input RGB image 
and the corresponding ground-truth rotation label belongs to the set of classes \( \mathcal{Y} = \{0, 1, 2, 3...,11\} \), where 0 corresponds to \( 0^\circ \), 1 to \( 30^\circ \), 2 to \( 60^\circ \), and so on, till \( 330^\circ \). To enhance robustness and model generalization, we adopt a \textit{dynamic cropping strategy}
~\cite{dong2024internvl}, where each input image \( x \) is transformed into multiple spatial crops \( \{x^{(1)}, x^{(2)}, \dots, x^{(K)}\} \). Each crop is independently passed through the vision encoder to produce a sequence of visual tokens $X^{(k)} \in \mathbb{R}^{L \times D}$ where \( D \) denotes embedding dimension, \( L \) and \( K \) the number of visual tokens and crops respectively. The vision encoder consists of a patch embedding module followed by a transformer encoder, consistent with the modern VLM architectures.
From each crop's output sequence, we extract the CLS token from the first position, \( h_{\text{cls}}^{(k)} = X_0^{(k)} \in \mathbb{R}^D \), which captures global contextual information, allowing to aggregate spatial and semantic cues for that crop. 
We compute the average CLS embedding over all \( K \) crops of the image to obtain a robust, unified image representation. 
We then use a classification head, which is a lightweight feedforward neural network \( f_{\text{clf}} \) that maps \( h_{\text{cls}} \) to one of 12 rotation classes:
\begin{equation}
\hat{y} = f_{\text{clf}}(h_{\text{cls}}) = W_2 \cdot \phi(W_1 \cdot h_{\text{cls}} + b_1) + b_2
\label{eq:classifier}
\end{equation}
where $W_1 \in \mathbb{R}^{D' \times D}$ and $b_1 \in \mathbb{R}^{D'}$ denote the weights and bias of the first layer, while $W_2 \in \mathbb{R}^{C \times D'}$ and $b_2 \in \mathbb{R}^{C}$ represent those of the output layer. We use the GELU activation~\citep{hendrycks2016gelu} function $\phi$, with $C$ denoting the number of rotation classes, and $D' = \frac{D}{2}$ specifying the hidden size of the intermediate layer. To mitigate overfitting, we applied dropout \citep{srivastava2014dropout} as $h_{\text{drop}} = \text{Dropout}(h_{\text{cls}})$ with a probability of 0.2. 
Thus, the complete forward pass of the classification head is:
\begin{equation}
\begin{aligned}
z &= \phi(W_1 \cdot h_{\text{drop}} + b_1) \in \mathbb{R}^{D'} \\
\hat{y} &= W_2 \cdot z + b_2 \in \mathbb{R}^C
\end{aligned}
\label{eq:forward-pass}
\end{equation}

The predicted logits \( \hat{y} \) are trained using softmax cross-entropy loss. For the 4-class setting, we set $C = 4$, while for fine-grained rotation classificaiton we use $C = 12$, without modifying the rest of the pipeline. This architecture enables compact and efficient inference while maintaining high classification accuracy. We adapt the vision encoder from the Phi-3.5 model~\cite{abdin2024phi3} as the backbone for our rotation classification task. By fine-tuning the pretrained 
vision encoder standalone on our proprietary document image datasets, we obtain a lightweight yet semantically powerful model that leverages the representational strength of CLIP ViT-L/14 model
while being optimized for rotation detection.
\section{OCR Rotation Benchmark (ORB)}
\label{sec:orb}

To address the lack of standardized benchmarks for rotation correction in OCR pipelines, we introduce OCR-Rotation-Bench (ORB), a new benchmark for evaluating OCR robustness to image rotations. ORB comprises two subsets: one for English-language documents (ORB-En) and another for multilingual Indic scripts (ORB-Indic). Both the datasets are \textit{evenly} distributed across either four or twelve rotation angles,
enabling balanced evaluation without any orientation bias.




\textbf{ORB-En:} 
We select three diverse and publicly available datasets to enable standardized, reproducible evaluation in the English OCR tasks:
\begin{itemize}[noitemsep, itemsep=0.5pt]
    \item SROIE 2019~\cite{huang2019sroie}, a widely-used benchmark comprising real-world scanned documents such as receipts and invoices, annotated at the \textit{field level} for structured OCR tasks.
    \item SynthDog~\cite{kim2022donut}, which contains \textit{free-form} synthetically generated English documents with complete OCR ground truth, offering controlled document structure and high annotation fidelity.
    \item FUNSD~\cite{jaume2019funsd}, a form understanding benchmark consisting of scanned forms with word-level text annotations, semantic entity labels, and explicit link annotations, enabling joint evaluation of text recognition and document layout understanding. 
\end{itemize}

We rotate the test and validation sets containing 347, 500 and 50 images, respectively of the three datasets \textit{uniformly} across the 4 or 12 rotation classes to simulate real-world scenarios such as upside-down or angled documents often seen in bulk-scanned OCR settings. This combination of real and synthetic data ensures that the benchmark is practically grounded and generalizable across OCR use-cases.

\begin{figure*}[htbp]
\centering
\setlength{\tabcolsep}{6pt}

\begin{tabular}{cc|cc}
\textbf{Multilingual Image} &
\makecell{\textbf{Rotation}\\\textbf{Classification}} &
\textbf{English Image} &
\makecell{\textbf{Rotation}\\\textbf{Classification}} \\
\hline

\includegraphics[width=0.28\textwidth]{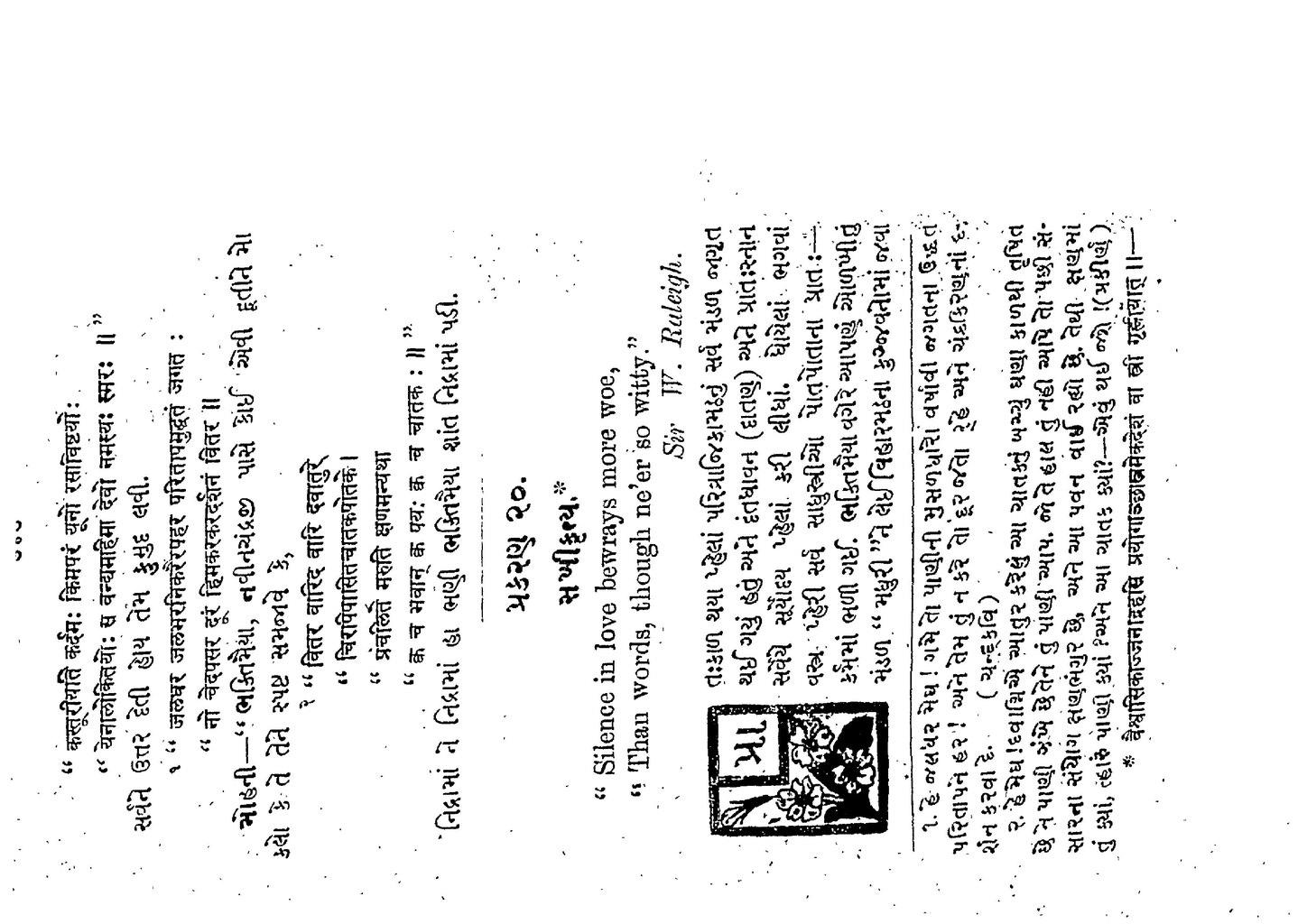} &
$270^\circ$ &
\includegraphics[width=0.28\textwidth]{sec/latex/resources/image/X51005337877.jpg} &
$0^\circ$ \\

\includegraphics[width=0.28\textwidth]{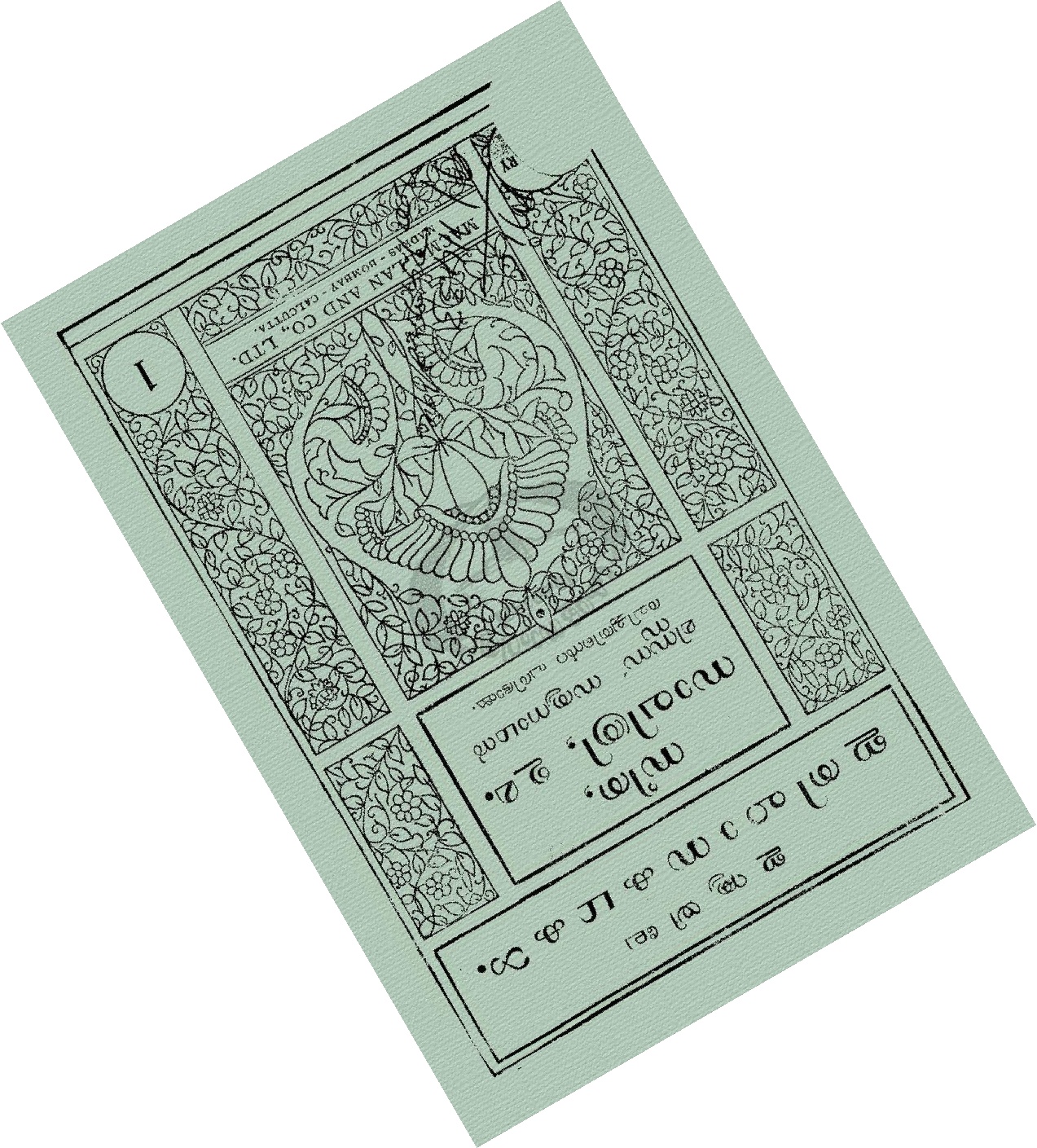} &
$150^\circ$ &
\includegraphics[width=0.28\textwidth]{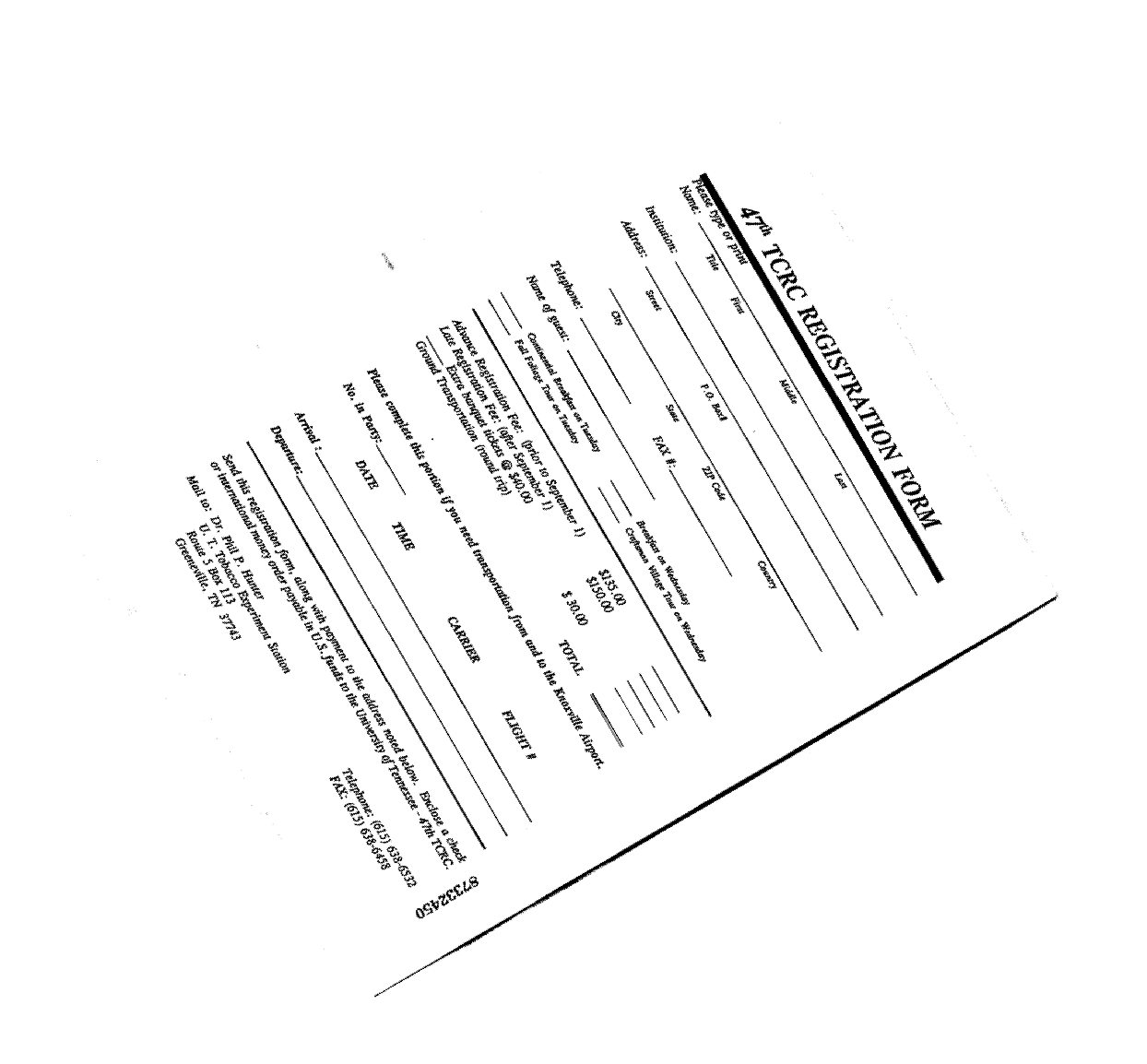} &
$60^\circ$ \\

\end{tabular}

\caption{Qualitative evaluation of rotation classification. Multilingual ORB-Indic samples and real-world documents are shown together, illustrating accurate orientation prediction and successful downstream OCR for both English and Indic languages.}
\label{fig:multilingual-rotation}

\end{figure*}










\textbf{ORB-Indic:} 
To construct the multilingual benchmark for rotation-aware OCR, we curate a high-quality Indic-language dataset from Wikisource, an open-access repository of digitized literary works. The dataset spans 11 Indic languages: Hindi, Bengali, Tamil, Malayalam, Telugu, Marathi, Kannada, Gujarati, Punjabi, Oriya, and Assamese, covering a wide typographic and linguistic diversity representative of real-world document understanding. 
Table \ref{tab:language-distribution} provides the balanced distribution of samples in each language. However, unlike concurrent work~\citep{anonymous2025multimodal}, our dataset additionally includes Assamese and specifically emphasizes robustness to rotation variations. We develop custom web crawlers to extract both document images and corresponding OCR text based on structured metadata provided by Wikisource. Specifically, 
\begin{itemize}[noitemsep]
    \item We filter for Level-4 verified pages, which are reviewed and annotated by human contributors on Wikisource, ensuring high textual fidelity in our dataset.
    \item Using the \texttt{prp\_page\_image} tag, we collect high-resolution page images.
    \item The corresponding OCR text is extracted via the \texttt{pagetext} tag, which reflects the best human-edited transcription.
\end{itemize}

To further ensure the quality and alignment of image-text pairs, we manually clean the collected data, discarding samples with mismatches, formatting issues, or incomplete annotations. The final test set comprises 966 document images, distributed across the 11 languages and twelve rotation classes. To the best of our knowledge, ORB-Indic is the first multilingual benchmark explicitly designed to evaluate the effect of rotation on OCR performance in Indic scripts. It provides a valuable resource for benchmarking and advancing rotation-robust OCR models in low-resource and multilingual settings.
\section{Experimental Design}
\label{sec:exp}

\begin{table}[htbp]
    \centering
    \begin{tcolorbox}
    \footnotesize

    \textbf{VLM Rotation Classifier prompt:} \\
    \texttt{Classify this image into one of 12 rotation classes. If the image is not rotated and upright, classify it as 0. If the image is rotated clockwise by 30 degrees, classify it as 30. If it is rotated clockwise by 60 degrees, classify it as 60. If it is rotated clockwise by 90 degrees, classify it as 90. If it is rotated clockwise by 120 degrees, classify it as 120. Continue similarly for 150, 180, 210, 240, 270, 300, and 330 degrees. Reply with just the class number (0, 30, 60, 90, 120, 150, 180, 210, 240, 270, 300, or 330) and nothing else.} 
    \\[1em]
    \textbf{English and Multilingual OCR prompt:} \\
    \texttt{You are an OCR engine. You job is to extract the exact text from the given image. Preserve the text formatting, including line breaks, spaces, and any symbols or special characters. Respond in plain text format.}
    
    \end{tcolorbox}
    \caption{Prompt templates used across tasks: We employ distinct prompt templates for (i) rotation classification via Vision Language Models (VLMs), and (ii) OCR in both English and Indic languages within our framework.}
    \label{tab:data-prompts}
\end{table}

In this section, we detail the experimental strategy used to evaluate both the intrinsic accuracy and downstream impact of the proposed rotation correction module. More technical details are provided in the Appendix \ref{appendix:implementation}. 

\textbf{Baselines:}
We benchmark against both conventional OCR models and 
Vision-Language Models (VLMs)
including
docTR~\cite{liao2023doctr},
transformer-based TrOCR~\cite{li2021trocr} and more recent VLMs including open-weights Gemma-3 (27B)~\cite{deepmind2024gemma}, Nanonets-OCR-s~\cite{nanonets2024ocr}, LLaMA-4-17B (Maverick) ~\cite{touvron2024llama4} supporting Indian languages, along with closed-source GPT-4o~\cite{hurst2024gpt4o} and Gemini-2.5 Pro and  Flash~\cite{team2023gemini}. All prompts are represented in Table~\ref{tab:data-prompts}. 


\textbf{Evaluation metrics:}
We report field-level accuracy on SROIE and free-form OCR on SynthDog under four rotation angles (0°, 90°, 180°, 270°). For ORB-Indic and the English FUNSD dataset, we report results under a 12-class rotation evaluation. We adopt the evaluation strategy used in prior and contemporary studies~\citep{fu2024ocrbench,anonymous2025multimodal}, reporting Average Normalized Levenshtein Similarity (ANLS) as an interpretable metric ~\citep{biten2019scene} in addition to other variants like Word Error Rate (WER) and Character Error Rate (CER)~\citep{smith2007overview,neudecker2021survey} based on Levenshtein Distance \citep{lcvenshtcin1966binary}.

\begin{table}[htbp]
\centering
\resizebox{\linewidth}{!}{%
\begin{tabular}{lccc}
\toprule
\textbf{Model} 
& \makecell[c]{\textbf{Ideal $\uparrow$}} 
& \makecell[c]{\textbf{w/o Rotation $\uparrow$}} 
& \makecell[c]{\textbf{w/ Rotation $\uparrow$}} \\
\midrule
Tesseract       & 49.06 & 24.06 & 48.99 \\
docTR           & 64.54 & 15.63 & 63.11 \\
\hdashline
Nanonets-OCR-s  & \underline{72.19} & \underline{64.27} & \underline{72.12} \\
Gemma-3 27B     & 68.37 & 53.24 & 68.23 \\
LLaMA-4 (Maverick) 17B & \underline{70.32} & 47.83 & \underline{70.32} \\
\hdashline
Qwen2.5-VL      & \textbf{72.62} & \underline{68.16} & \textbf{72.26} \\
mPLUG-DocOwl2   & 7.06  & 2.67  & 6.77 \\
\hdashline
GPT-4o          & 36.09 & 29.10 & 35.87 \\
Gemini-2.5 Pro  & 62.82 & 46.61 & 60.66 \\
Gemini-2.5 Flash & 70.10 & \textbf{69.67} & 70.10 \\
\bottomrule
\end{tabular}}
\caption{Field-level OCR accuracy (in \%) on ORB-En-SROIE with and w/o our rotation pipeline.
We observe substantial gains for classical and open-source OCR engines, while large VLMs exhibit mixed robustness to rotation.}
\label{tab:ocr-comparison}
\end{table}

\textbf{Impact on Downstream OCR:} We evaluate the downstream OCR performance of all baselines under a controlled setup, considering three scenarios:

\begin{itemize}[noitemsep, itemsep=0.5pt]
\item \textit{Ideal Upright Conditions:} 
OCR on upright images with an ideal canonical setup.
\item \textit{Without Pipeline:} OCR on randomly rotated images, resembling orientation patterns seen in practical scenarios.
\item \textit{With Pipeline:} OCR after orientation normalization where images are realigned using our rotation classification network. Any misclassifications by the rotation model may introduce deviations from the ideal upright condition, potentially impacting downstream OCR performance of the baseline models.
\end{itemize}
\section{Results and Discussion}
\label{sec:results}
We summarize the key experimental findings here of the rotation classification as well as the downstream performance of considered models.

\begin{table}[htbp]
\small
\centering
\setlength{\tabcolsep}{6pt}      
\renewcommand{\arraystretch}{1.1}
\begin{tabular}{lcc}
\toprule
\textbf{Model} 
& \textbf{ORB-En} 
& \textbf{ORB-Indic} \\
\midrule
TrOCR 
& \underline{78.18} & \underline{76.29} \\

docTR 
& 10.00 & 17.39 \\
\hdashline

Nanonets-OCR-s 
& 8.00 & 21.53 \\

Gemma-3 27B 
& 16.00 & 18.74 \\

LLaMA-4 (Maverick) 17B 
& 32.00 & 29.19 \\
\hdashline

GPT-4o 
& 34.00 & 17.18 \\

Gemini-2.5 Flash 
& 22.00 & 23.63 \\

Gemini-2.5 Pro 
& 24.00 & 22.05 \\
\hdashline

\textbf{Ours} 
& \textbf{98.00} & \textbf{96.71} \\
\bottomrule
\end{tabular}
\caption{Accuracy (\%) for 12-class document rotation classification on ORB-En and ORB-Indic. Best in bold, second-best underlined. Conventional models like TrOCR and docTR were finetuned for rotation classification while others are used in 0-shot setting.}
\label{tab:12class-accuracy}
\end{table}

\textbf{Rotation Classification:} 

Table~\ref{tab:12class-accuracy} and  Table~\ref{tab:4class-accuracy} (Section \ref{appendix:ablation} in Appendix) report the rotation classification accuracy of all evaluated models on our dataset on 2 different cases: 4 class canonical angles (0, 90, 180 and 270 degree rotations) and 12 class rotation problem (0, 30, 60, 90, 120, 150, 180, 210, 240, 270, 300, 330 degree angles). Our proposed classifier, comprising approximately 304M parameters, with 0.5s latency on NVIDIA H100, achieves 96.81\% accuracy on ORB-En and 92.68\% on the multilingual ORB-Indic dataset on the 4-class classification problem and 98\% on ORB-En and 96.71\% on ORB-Indic on the 12-class classification problem respectively, demonstrating strong generalization across diverse scripts and layouts. We also conducted 2 ablation studies where dynamic cropping with multi-layer classifier showed the best performance balancing both accuracy and latency shown in Table~\ref{tab:ablation-study} and Table~\ref{tab:ablation-dynamic-crop-funsd} (Section \ref{appendix:ablation} in Appendix).

Conventional OCR-based systems such as docTR and TrOCR achieve the second-best performance, particularly on the more challenging 12-class rotation task, after being fine-tuned for rotation classification. In contrast, all evaluated VLMs perform poorly, corroborating recent evidence that VLMs struggle on standard computer vision benchmarks~\citep{zhang2024visually, cooper2025rethinking, ramachandran2025well}. Models such as Gemma-3, LLaMA-4, GPT-4o, and Gemini-2.5 fail to reliably infer document orientation, with GPT-4o marginally outperforming Gemini-2.5 Flash among closed-source models, yet remaining far below our approach. A classical, training-free baseline based on HOG and SIFT orientation statistics yields near-zero exact-match accuracy under the 12-class (30°) setting for both English and Indic datasets, underscoring the limitations of handcrafted methods at fine-grained rotation resolution.

\textbf{Evaluation on structured OCR:} We evaluate the effect of our rotation correction module on downstream OCR using ORB-En-SROIE (Table~\ref{tab:ocr-comparison}) under the 4-class setting. Field-level accuracy is computed over four equally weighted fields (25\% each). Rotation correction yields substantial gains for classical and open-source OCR systems: Tesseract nearly doubles its accuracy, while docTR improves by almost fourfold, underscoring their strong sensitivity to orientation errors. Nanonets-OCR-s attains strong absolute performance and benefits modestly from correction, indicating limited inherent rotation robustness.

Vision–language models show mixed trends. Qwen2.5-VL achieves the best overall accuracy under ideal and corrected settings, with only marginal gains from explicit correction, suggesting partial rotation robustness. In contrast, mPLUG-DocOwl2 performs poorly across all settings. Among closed-source models, Gemini-2.5 Pro benefits noticeably from correction, whereas Gemini-2.5 Flash shows minimal improvement, likely due to internal rotation handling, and GPT-4o remains inferior to conventional OCR engines on structured extraction.

All results are reported without task-specific fine-tuning, highlighting the effectiveness of a lightweight standalone rotation correction module across diverse OCR paradigms.

\begin{table*}[htbp]
\centering
\resizebox{\textwidth}{!}{%
\begin{tabular}{lcccccc}
\toprule
\textbf{Model} 
& \multicolumn{2}{c}{\textbf{Ideal upright conditions}} 
& \multicolumn{2}{c}{\textbf{w/o Rotation}} 
& \multicolumn{2}{c}{\textbf{w/ Rotation}} \\
\cmidrule(lr){2-3} \cmidrule(lr){4-5} \cmidrule(lr){6-7}
& \textbf{Word $\downarrow$} & \textbf{Character $\downarrow$} 
& \textbf{Word $\downarrow$} & \textbf{Character $\downarrow$} 
& \textbf{Word $\downarrow$} & \textbf{Character $\downarrow$} \\
\midrule

\textit{ORB-En-FUNSD dataset} \\
docTR  & 38.37 & \underline{15.02} & 93.48 & 75.97 & 39.86 & \textbf{16.60} \\
\hdashline
Nanonets-OCR-s       & 99.38 & 99.23 & 99.58 & 99.23 & 99.41 & 99.21 \\
Gemma-3 27B          & 37.41 & 25.57 & 73.81 & 60.36 & 38.84 & 26.97 \\
LLaMA-4 (Maverick) 17B & 54.44 & 32.43 & 83.46 & 67.18 & 54.98 & 33.82 \\
\hdashline
Qwen2.5-VL           & \textbf{31.66} & 16.72 & 63.68 & \underline{51.28} & \textbf{33.08} & \underline{18.20} \\
\hdashline
GPT-4o               & 38.05 & 37.98 & 79.91 & 72.68 & 38.05 & 37.98 \\
Gemini-2.5 Pro       & 46.49 & 35.82 & \underline{62.81} & \underline{51.44} & 47.23 & 36.53 \\
Gemini-2.5 Flash     & \underline{34.98} & \textbf{11.93} & \textbf{53.68} & \textbf{39.33} & \underline{33.26} & 19.42 \\
\midrule

\textit{ORB-Indic dataset} \\
Nanonets-OCR-s       & 91.68 & 86.92 & 97.21 & 95.58 & 91.68 & 86.92 \\
Gemma-3 27B          & 70.38 & 45.47 & 95.67 & 80.50 & 71.61 & 47.29 \\
LLaMA-4 (Maverick) 17B & 46.68 & 23.88 & 92.99 & 79.50 & \underline{47.01} & \underline{24.56} \\
\hdashline
GPT-4o               & 83.36 & 65.51 & 97.70 & 93.53 & 83.25 & 65.79 \\
Chitrapathak         & \textbf{34.05} & \textbf{15.81} & 93.53 & 80.40 & \textbf{35.43} & \textbf{17.57} \\
Gemini-2.5 Pro       & 41.97 & 27.26 & \underline{81.49} & \underline{71.67} & 51.51 & 37.83 \\
Gemini-2.5 Flash     & \underline{35.02} & \underline{16.30} & \textbf{66.36} & \textbf{49.53} & 53.66 & 36.27 \\

\bottomrule
\end{tabular}}
\caption{ANLS (Average Normalized Levenshtein Similarity) under \textbf{12-class rotation} across English and Indic datasets for different models. Word- and character-level scores are reported for ideal upright inputs, OCR without rotation correction, and OCR with rotation correction (lower is better).}
\label{tab:merged-ocr-anls-12class}
\end{table*}

\textbf{Evaluation on free-form OCR:}
Table~\ref{tab:merged-ocr-anls-12class} and Table~\ref{tab:merged-ocr-anls-4class} report word- and character-level Average Normalized Levenshtein Similarity (ANLS; lower is better) across English and Indic benchmarks under three settings: ideal upright documents, rotated documents without correction, and rotated documents with rotation correction, for both 4-class and 12-class rotation problems.

On \textit{ORB-En-SynthDog} (4-class), most OCR systems and VLMs exhibit strong orientation sensitivity. docTR and Gemma-3 degrade sharply under rotation, with word-level ANLS increasing from 56.2 to 88.0 and from 61.6 to 75.2, respectively. Nanonets-OCR-s, despite near-optimal upright performance, degrades severely under rotation (2.6 to 44.1), indicating pronounced orientation bias~\citep{roberts2023data,deng2024unveiling}. VLM robustness is mixed: Gemini-2.5 Flash remains largely invariant (22.4 to 26.2) and performs best without correction, Gemini-2.5 Pro benefits moderately from correction, Qwen-2.5-VL shows partial gains, and mPLUG-DocOwl2 performs poorly across all settings. TrOCR and Tesseract are omitted as unsuitable for free-form documents.

The challenge is further amplified on \textit{ORB-En-FUNSD} (12-class), which introduces finer-grained orientation variability on unseen documents. All models degrade without correction; however, Gemini-2.5 Flash attains the lowest ANLS under rotation correction, reducing word-level error by nearly 20 points (53.68→33.26). Qwen-2.5-VL shows competitive gains but remains inferior to Gemini-2.5 Flash. docTR, Gemma-3, and LLaMA-4 exhibit compounded errors, underscoring the difficulty of fine-grained orientation handling for end-to-end OCR systems.

On the multilingual \textit{ORB-Indic} benchmark (966 images across 11 Indic languages), fine-grained orientation effects are amplified in the 12-class setting. Under upright conditions, Chitrapathak achieves the lowest ANLS (34.05 / 15.81), followed closely by Gemini-2.5 Flash (35.02 / 16.30), establishing them as the strongest performers. In contrast, general-purpose OCR and VLM-based systems incur substantially higher errors. Notably, Gemini-2.5 Flash remains the most robust under uncorrected rotation (66.36 / 49.53), despite an almost twofold increase in word-level ANLS. Overall, these results expose a clear gap between Indic-specialized and general-purpose models under fine-grained rotation and highlight the difficulty of multilingual OCR without explicit orientation handling.

Critically, explicit rotation correction is far more effective in the 12-class setting. While several models exhibit partial invariance to coarse (4-class) rotations, fine-grained orientation variability causes consistent degradation across all OCR pipelines when correction is omitted. Incorporating our lightweight rotation module yields substantially larger gains than in the 4-class case, particularly on ORB-En-FUNSD and ORB-Indic (12-class), where models recover a significant portion of upright performance.

Overall, these results indicate that implicit orientation robustness in large OCR and vision–language models is insufficient under fine-grained rotations. Decoupling orientation estimation from recognition consistently improves downstream OCR accuracy, especially for multilingual and free-form documents where visual and linguistic complexity amplifies orientation sensitivity. Additional WER and CER results are reported in Table~\ref{tab:merged-ocr-comparison-hf} (Appendix), with percentage-based similarity scores in Table~\ref{tab:merged-ocr-comparison}. Across all settings, Gemini-2.5 Flash remains the most rotation-robust OCR model on both English and Indic benchmarks.

\textbf{Ablation study:} 
We also performed an ablation study analyzing three variants of our model in Table~\ref{tab:ablation-study}:

\begin{itemize}
    \item \textit{Single-layer classifier:} Single linear layer on encoder output, with dynamic cropping.
    \item \textit{Multi-layer classifier:} Deep head with GELU and Dropout, without dynamic cropping.
    \item \textit{Ours (Final model):} Multi-layer classifier with dynamic cropping and multi-level classifier head.
\end{itemize}

Our ablation study highlights the individual contributions of dynamic cropping and classifier complexity to overall model performance. The baseline variant with a single-layer classifier and dynamic cropping achieves moderate accuracy, indicating that while spatial diversity in input patches is beneficial, a shallow classification head may under utilize the rich representations from the Phi-3.5 encoder. Replacing it with a deeper multi-layer classifier improves results across both ORB-English and ORB-Indic datasets, even without cropping, suggesting that classifier depth significantly enhances discriminative capacity. 

Finally, combining both techniques i.e., dynamic cropping and a multi-layer classifier achieves the highest accuracy on both benchmarks (96.81\% on ORB-En and 92.68\% on ORB-Indic) on the 4-class problem, demonstrating that dynamic cropping provides complementary spatial context that further boosts performance when paired with a sufficiently expressive classifier. This underscores the importance of architectural synergy in maximizing the rotation classification capability of vision-language models.


\begin{table}[htbp]
\centering
\resizebox{0.9\linewidth}{!}{%

\begin{tabular}{p{0.3\linewidth}c@{\hspace{0.5em}}c}
\toprule
\textbf{Model Variant} & \textbf{ORB-En (\%)} & \textbf{ORB-Indic (\%)} \\
\midrule
\makecell[l]{Single-layer classifier\\(w/ dynamic cropping)} & 90.66 & 82.01 \\
\makecell[l]{Multi-layer classifier\\(w/o dynamic cropping)} & 92.65 & 86.59 \\
\makecell[l]{\textbf{Multi-layer classifier}\\\textbf{with dynamic cropping}\\\textbf{(Ours)}} & \textbf{96.81} & \textbf{92.68} \\
\bottomrule
\end{tabular}
}
\caption{Ablation results comparing classifier complexity and dynamic cropping on ORB-English and ORB-Indic for 4 class classification problem.}
\label{tab:ablation-study}
\end{table}

\begin{figure}[htbp]
\centering
\includegraphics[width=\linewidth]{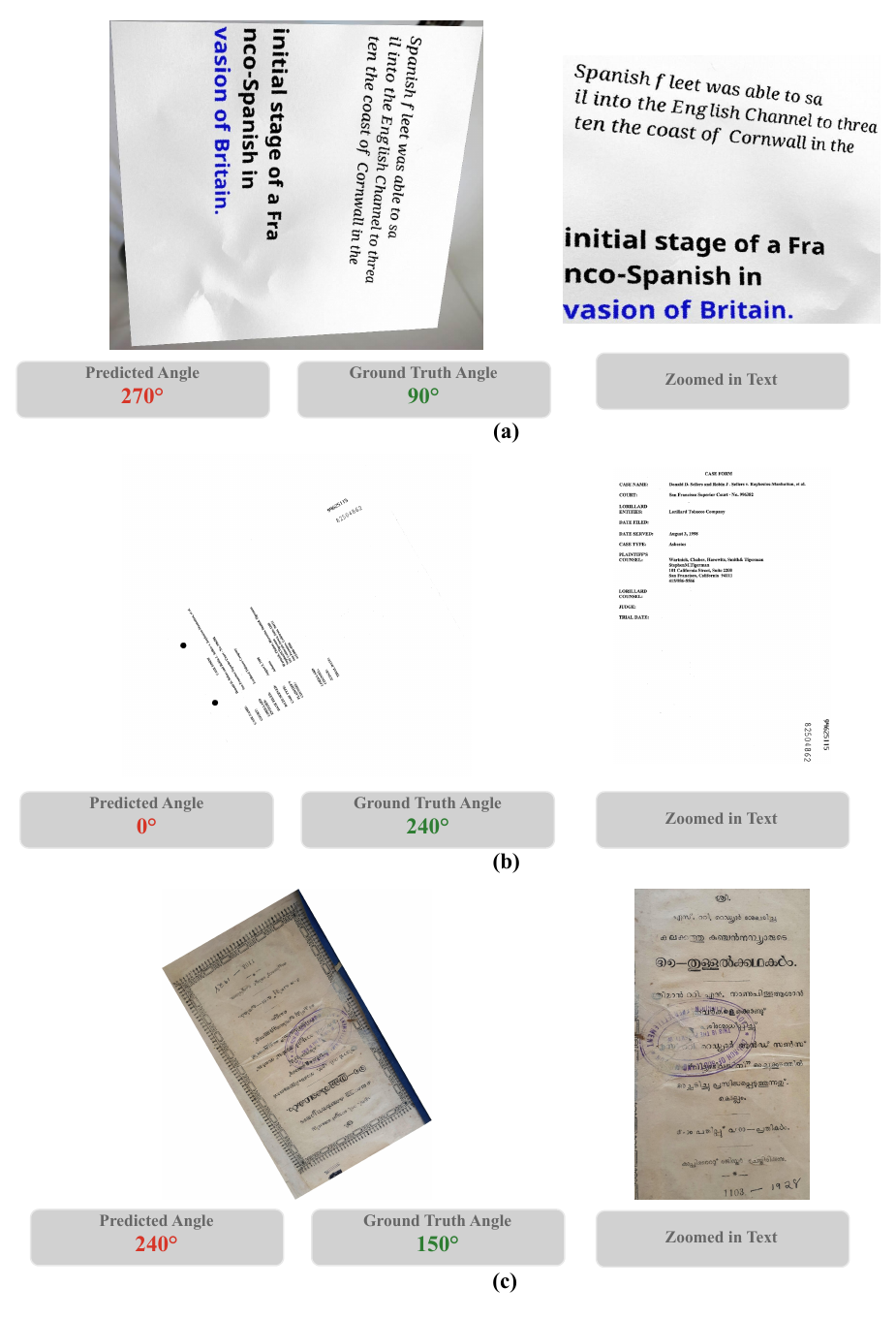}
\caption{We show failure cases from the ORB-En-SynthDog (a), ORB-En-FUNSD (b) and ORB-Indic datasets (c), discussed in Section \ref{sec:results}.}
\label{fig:failure-cases}
\end{figure}

\textbf{Failure Analysis:} We manually inspect failure cases of our rotation module across both ORB-En and ORB-Indic benchmark datasets. In the ORB-Indic test set, 45 images, mostly in Tamil, Hindi, and Telugu were misclassified, primarily confusing $90^\circ$, $0^\circ$ and $240^\circ$ with $270^\circ$, $180^\circ$ or $120^\circ$, respectively. 
These errors were typically due to the text being not a majority of the document as well as non uniform placement of the text in the document. 

In contrast, only 2 misclassifications occurred in the ORB-En-SROIE and 1 in ORB-En-FUNSD, where $90^\circ$ and $240^\circ$ samples were predicted as $270^\circ$ and $0^\circ$. This is primarily due to excessive padding around the text regions as well as off-centered text. However, we observed 25 similar misclassifications in the ORB-En-SynthDog subset, indicating a higher sensitivity to synthetic document variability as well as skewed placement of the text in the document.  
Figure~\ref{fig:failure-cases} presents visual examples of these non-standard document layouts, highlighting model confusion caused by unbalanced margins, uneven or extended padding around text regions.

\section{Conclusion}
\label{sec:conclusion}
This work highlights the importance of rotation correction in OCR pipelines, demonstrating up to 4× improvement in traditional OCR systems. To rigorously assess models, we introduce OCR-Rotation-Bench (ORB) with 1863 images, comprising (i) ORB-En, derived from rotation-transformed English datasets (structured and free-form), and (ii) ORB-Indic, a multilingual set of real-world documents across 11 Indic languages with ground-truth annotations. We present a lightweight rotation classifier built on the Phi-3.5 vision encoder, achieving near-perfect accuracy. Future extensions include arbitrary-angle rotation, additional Indic documents, a unified multilingual model, and layout-aware OCR for handwritten and irregular documents.

Our evaluations reveal significant limitations in SOTA VLMs on basic vision tasks like rotation classification, despite their broader capabilities. These results suggest that specialized lightweight models can remain highly competitive for fundamental document preprocessing tasks. Furthermore, our findings highlight the importance of benchmarking simple yet critical tasks that are often overlooked in large-scale multimodal model evaluations. Ultimately, our findings underscore the need for future progress grounded in well-defined tasks, realistic datasets, and rigorous evaluation protocols for Document Intelligence. We hope ORB and our benchmarking protocol will serve as a foundation for future research on robust document orientation and preprocessing in multilingual OCR pipelines.

{
    \small
    \bibliographystyle{ieeenat_fullname}
    \bibliography{main}
}

\clearpage
\setcounter{page}{1}
\maketitlesupplementary

\section{Ethics Statement}

This work centers on the responsible creation of a benchmark for evaluating OCR robustness to rotation across English and Indic languages. All data used in this study originates from publicly available sources such as Wikisource and established OCR datasets (e.g., SROIE, SynthDog, FUNSD). We ensured that no sensitive or personally identifiable information (PII) was collected or included. The Indic subset, derived from Wikisource, contains only publicly released literary and historical documents that are already digitized and verified. As the research involves only publicly available and non-personal data, formal IRB approval was not required. All external datasets, libraries, and tools are appropriately credited through citations. We exercised care to ensure fairness across scripts and languages, though residual bias may persist due to unequal digital representation among Indic languages. The resulting OCR-Rotation-Bench (ORB) aims to support the development of robust, multilingual, and rotation-aware OCR systems for broader real-world document understanding.

\section{Reproducibility Statement}

The full training pipeline, inference scripts, and our benchmark ORB (OCR-Rotation-Bench), and pretrained weights for both English and Indic datasets will be made available upon acceptance to ensure reproducibility. Our experimental settings and evaluation protocols are comprehensively documented to enable exact replication of results.

\section{Implementation Details}
\label{appendix:implementation}

\subsection{Model Architecture}



We adapt the image encoder from the Phi-3.5 model~\cite{abdin2024phi3} as the backbone for our rotation classification task. To adapt the encoder for classification, we append a lightweight two-layer classification head consisting of a linear projection, followed by a GELU activation, dropout, and a final linear layer. This classification head operates on a pooled image representation and outputs logits over 12 rotation classes: 0°, 30°, 60°and so on, till 330°.

Prior to encoding, input images undergo a dynamic cropping strategy inspired by the InternVL2-style sampling method~\cite{internvl2_github}. Specifically, each image is first scaled and padded to the closest HD resolution such that both height and width are divisible by 336. The padded image is then divided into non-overlapping crops of size 336×336, with a maximum of 16 crops per image. In addition to these local crops, we include a globally resized version of the full image (also 336×336) to capture holistic context. Each of these crops, including the resized full image, is independently passed through the image encoder, yielding 577 tokens (based on $14 \times 14$ patch sizes) per crop, including the [CLS] token, each with a hidden dimension of 1024. We extract the [CLS] token from each encoded crop and compute their mean across all crops and the full image. This averaged [CLS] representation is then used as input to the classification head.

\subsection{Implementation Details}

We performed all experiments in PyTorch~\citep{paszke2019pytorch} using HuggingFace Transformers~\citep{wolf2019huggingface} library (v4.41.2). Training employs the AdamW~\citep{kingma2020method} optimizer with a constant learning rate of $1 \times 10^{-5}$, and cross-entropy loss. We used a batch size of 256 for variants without dynamic cropping, and 64 for those with dynamic cropping. To improve stability, gradient clipping with a maximum norm of 1.0 was applied. Each model is trained for up to 45 epochs with early stopping based on validation accuracy. 
We use a lightweight classification head (two fully connected layers with non-linear activations) to predict one of twelve rotation classes. 
All training was performed on a single NVIDIA H100 GPU, which has an average latency of 0.5s per image during inference. The same model has a latency of around 0.9s on a single NVIDIA A100 GPU. We used the implementation of OCRBenchV2\footnote{\url{https://github.com/Yuliang-Liu/MultimodalOCR/blob/main/OCRBench_v2/eval_scripts/vqa_metric.py}} to compute the ANLS metrics. We adapted this script to report both word-level and character-level ANLS metrics. For structured analysis on ORB-SROIE, we test the model against 4 fields: company, date, address and total for the percentage match calculations. We provide more details about the configurations used across different baselines.





\paragraph{Tesseract \cite{smith2007overview}} 
For the \textit{Tesseract} baseline, we utilized the open-source \texttt{pytesseract} interface (version~5.3.0) built on top of the Tesseract OCR engine~(v5.2). Each image was converted to RGB and processed using the \texttt{pytesseract.image\_to\_string()} method to extract text content. 
The extracted text was lowercased and evaluated at the field level by matching against ground-truth JSON annotations.

\begin{itemize}[noitemsep, itemsep=0.5pt]
    \item \textbf{Model:} Tesseract OCR Engine (v5.2)
    \item \textbf{Interface:} \texttt{pytesseract} (v5.3.0)
    \item \textbf{Language model:} English (default)
    \item \textbf{Preprocessing:} RGB conversion using PIL
    \item \textbf{Framework:} Python~3.10
    \item \textbf{Temperature:} N/A (non-generative)
    \item \textbf{Max tokens:} N/A
    \item \textbf{Decoding:} Greedy text extraction
\end{itemize}

\paragraph{TrOCR \cite{li2021trocr}} 
For the \textit{TrOCR} baseline, we used the transformer-based \texttt{microsoft/trocr-base-printed} model consisting of a Vision Transformer (ViT) encoder and a GPT-2 style decoder. The model was loaded via \texttt{VisionEncoderDecoderModel} with the corresponding \texttt{TrOCRProcessor} for preprocessing and tokenization. Each image was resized to $384{\times}384$, and text was generated autoregressively using beam search decoding. Outputs were lowercased before evaluation against field-level annotations.

\begin{itemize}[noitemsep, itemsep=0.5pt]
    \item \textbf{Model:} \texttt{microsoft/trocr-base-printed}
    \item \textbf{Architecture:} Vision Transformer (encoder) + GPT-2 decoder
    \item \textbf{Framework:} PyTorch (transformersv4.41)
    \item \textbf{Processor:} \texttt{TrOCRProcessor}
    \item \textbf{Resolution:} $384{\times}384$
    \item \textbf{Max length:} 512 tokens
    \item \textbf{Decoding:} Beam search
    \item \textbf{Temperature:} 1.0 (default)
\end{itemize}

\paragraph{docTR \cite{liao2023doctr}} 
For the \textit{docTR} baseline, we employed the open-source \href{https://mindee.github.io/doctr/}{docTR} framework using the built-in \texttt{ocr\_predictor()} function with a MobileNetV3 backbone for both detection and recognition. Specifically, we used the \texttt{db\_mobilenet\_v3\_large} detector and \texttt{crnn\_mobilenet\_v3\_small} recognizer, both initialized with pretrained weights. The pipeline performs text detection using Differentiable Binarization (DB) and recognition using a convolutional recurrent network (CRNN). 

\begin{itemize}[noitemsep, itemsep=0.5pt]
    \item \textbf{Detector:} \texttt{db\_mobilenet\_v3\_large}
    \item \textbf{Recognizer:} \texttt{crnn\_mobilenet\_v3\_small}
    \item \textbf{Framework:} docTR~1.6.0
    \item \textbf{Backend:} PyTorch (CPU/GPU)
    \item \textbf{Temperature:} N/A (non-generative)
    \item \textbf{Max tokens:} N/A
    \item \textbf{Decoding:} Greedy decoding
    \item \textbf{Batch size:} 1
\end{itemize}

\paragraph{Nanonets-OCR-s \cite{nanonets2024ocr}} 
For the \textit{Nanonets-OCR-s} baseline, we employed the multimodal model released under the \texttt{nanonets/Nanonets-OCR-s} repository via the Hugging~Face \texttt{transformers} interface. The model is built upon the \texttt{Qwen2.5-VL} architecture for vision-language understanding and image-to-text generation. Inference was executed using \texttt{flash\_attention\_2} with automatic device mapping and mixed precision. Generation was performed deterministically without sampling. The \texttt{AutoProcessor} handled image and text preprocessing, while outputs were decoded using the corresponding \texttt{AutoTokenizer}.

\begin{itemize}[noitemsep, itemsep=0.5pt]
    \item \textbf{Model:} \texttt{nanonets/Nanonets-OCR-s}
    \item \textbf{Architecture:} Qwen2.5-VL (vision-language encoder–decoder)
    \item \textbf{Framework:} Hugging~Face Transformers (v4.41)
    \item \textbf{Attention:} \texttt{flash\_attention\_2}
    \item \textbf{Max new tokens:} 15000
    \item \textbf{Temperature:} 0.0
\end{itemize}

\paragraph{Gemma-3 27B \cite{team2024gemma}} 
For the \textit{Gemma-3 27B} baseline, we used the multimodal instruction-tuned \texttt{Gemma-3-27B-IT} model served through an API endpoint. Each request included a textual instruction and an inline Base64-encoded image. Responses were obtained in deterministic mode with default decoding. All predictions were saved to a CSV file for downstream evaluation.

\begin{itemize}[noitemsep, itemsep=0.5pt]
    \item \textbf{Model:} \texttt{Gemma-3-27B-IT}
    \item \textbf{Max tokens:} Default 
    \item \textbf{Temperature:} 0.0 
    \item \textbf{Decoding:} Greedy text generation
\end{itemize}

\paragraph{GPT-4o \cite{hurst2024gpt4o}} 
For the \textit{GPT-4o} baseline, we used OpenAI’s multimodal \texttt{gpt-4o} model through an API endpoint. Images were encoded as Base64 inline URIs and passed via the \texttt{image\_url} message field. Inference was deterministic with no sampling, and the model returned a single plain-text OCR response per image.

\begin{itemize}[noitemsep, itemsep=0.5pt]
    \item \textbf{Model:} \texttt{gpt-4o}
    \item \textbf{Provider:} OpenAI
    \item \textbf{Input format:} Base64-encoded inline image
    \item \textbf{Max tokens:} 2048
    \item \textbf{Temperature:} 0.0 
    \item \textbf{Decoding:} Greedy text generation
    \item \textbf{Response type:} Plain text OCR output
    \item \textbf{Framework:} \texttt{openai} Python SDK (v1.35)
\end{itemize}

\paragraph{LLaMA-4 (Maverick) 17B \cite{touvron2024llama4}} 
For the \textit{LLaMA-4 (Maverick) 17B} baseline, we utilized the multimodal instruction-tuned model \texttt{Llama-4-Maverick-17B-128E-Instruct}.
The model received both the image and the instruction prompt and produced plain-text OCR outputs deterministically.

\begin{itemize}[noitemsep, itemsep=0.5pt]
    \item \textbf{Model:} Llama-4-Maverick-17B-128E-Instruct
    \item \textbf{Max tokens:} 2048
    \item \textbf{Temperature:} 0.0 
    \item \textbf{Decoding:} Greedy text generation
\end{itemize}

\paragraph{Chitrapathak} 
For the \textit{Chitrapathak} baseline, we employed a multilingual vision–language OCR model hosted via the API endpoint\footnote{\url{https://bit.ly/chitrapathak}}. 
The model returned plain-text OCR predictions, which were parsed and saved incrementally to CSV files.

\begin{itemize}[noitemsep, itemsep=0.5pt]
    \item \textbf{Model:} \texttt{chitrapathak}
    \item \textbf{Prompt:} Implicit OCR generation (no explicit instruction)
    \item \textbf{Decoding:} Greedy text generation
    \item \textbf{Response type:} Plain text OCR output
\end{itemize}

\paragraph{Gemini-2.5 Pro \cite{google2025gemini2.5pro}} 
For the \textit{Gemini-2.5 Pro} baseline, we used the multimodal version of Gemini-2.5 model. Each image was encoded as a Base64 inline image and passed along with the instruction prompt. 
Requests were made
with a maximum token limit of 2048 and a low sampling temperature to ensure deterministic decoding. The returned text was stored directly as the OCR output in CSV format.

\begin{itemize}[noitemsep, itemsep=0.5pt]
    \item \textbf{Model:} \texttt{google/gemini-2.5-pro}
    \item \textbf{Access:} \texttt{google.genai} SDK
    \item \textbf{Input format:} Base64-encoded inline image
    \item \textbf{Max tokens:} 2048
    \item \textbf{Temperature:} 0.1
    \item \textbf{Decoding:} Greedy text generation
    \item \textbf{Response type:} Plain text OCR output
\end{itemize}

\paragraph{Gemini-2.5 Flash \cite{google2025gemini2.5flash}} 
For the \textit{Gemini-2.5 Flash} baseline, we used Google’s multimodal model accessed via the \texttt{google.genai} Python SDK. The model \texttt{gemini-2.5-flash} was configured with an authenticated API key and executed in streaming mode for OCR text extraction. Each image was read as binary data and sent to the model as an inline MIME-typed input. The outputs were returned as plain-text responses and stored incrementally in JSONL format for further evaluation.

\begin{itemize}[noitemsep, itemsep=0.5pt]
    \item \textbf{Model:} \texttt{google/gemini-2.5-flash}
    \item \textbf{Access:} \texttt{google.genai} SDK
    \item \textbf{Input format:} Binary image with MIME type (\texttt{image/jpeg}, \texttt{image/png}, \texttt{image/webp})
    \item \textbf{Response format:} Plain text
    \item \textbf{Temperature:} Default (deterministic)
    \item \textbf{Decoding:} Greedy text generation
    \item \textbf{Output storage:} JSONL file
\end{itemize}





\subsection{Data Sources and Augmentation}

For English training data, we use documents from various categories such as invoices, rent agreements, electricity bills, and gas bills, collected from internal repositories. For Indic data, we curate document images from 11 languages (Hindi, Bengali, Tamil, Malayalam, Telugu, Marathi, Kannada, Gujarati, Punjabi, Oriya, and Assamese) using Wikisource. To ensure fair evaluation, we strictly exclude all test images from the training sets. Specifically, the ORB-En and ORB-Indic benchmarks are fully disjoint from the training data. Moreover, we do not use the SROIE 2019 or the FUNSD training split during model training. All results reported are therefore in a zero-shot setting, highlighting the generalization capability of the proposed method across both English and Indic scripts on our benchmark.

Rotation transformation was applied uniformly across all four rotation classes. The proprietary English dataset used for training and validation comprised approximately 11K document images. For multilingual training, we used around 38K images spanning 11 Indic scripts in our dataset. 

For ORB-Indic, our data collection process began with the official Wikimedia XML snapshots\footnote{\url{https://dumps.wikimedia.org/hiwikisource/latest/}}, obtained from the official Wikimedia snapshots \footnote{\url{https://dumps.wikimedia.org/hiwikisource/latest/}}. We parsed the extracted XML files to identify individual page titles and systematically reconstructed their canonical Wikisource URLs for downstream processing. We curated high-quality page images and manually verified ground-truth orientation. Image rotation was done using the Pillow\footnote{\url{https://pypi.org/project/pillow/}} and OpenCV\footnote{\url{https://opencv.org/}} libraries while ensuring that the data covered all 4 classes equally.

\subsection{Evaluation and Baselines}
\label{appendix:evaluate-wer}

We use HuggingFace evaluate \footnote{\url{https://huggingface.co/spaces/evaluate-metric/wer/blob/main/wer.py}} library to also compute the WER and CER metrics in Table \ref{tab:merged-ocr-comparison-hf}. In addition, we also report a modified version using the \texttt{SequenceMatcher} utility from Python's \texttt{difflib} module in Table \ref{tab:merged-ocr-comparison}, however we follow the ANLS metric for its simplicity and interpretation.  
We will release all the evaluation scripts upon acceptance. We evaluate our method on the curated benchmark datasets:


\begin{enumerate}
    \item \textbf{ORB-En:} A 897-image benchmark composed of 347 images from the SROIE 2019 test set, 500 images from the SynthDog validation set and 50 images from the FUNSD validation set, uniformly rotated and re-annotated.
    \item \textbf{ORB-Indic:} A 966-image benchmark covering 11 Indic scripts across 12 different angles, each manually annotated and verified for rotation labels, along with corresponding ground-truth transcriptions for OCR evaluation.
\end{enumerate}

We compare against several strong baselines, including both traditional and VLM-based OCR models:

\begin{itemize}[noitemsep, itemsep=0.5pt]
    \item \textbf{Tesseract}~\cite{smith2007overview}
    \item \textbf{TrOCR}~\cite{li2021trocr}
    \item \textbf{docTR}~\cite{liao2023doctr}    
    \item \textbf{Nanonets-OCR-s}~\cite{nanonets2024ocr}
    \item \textbf{Gemma-3 27B}~\cite{team2024gemma}
    \item \textbf{LLaMA-4 (Maverick) 17B}~\cite{touvron2024llama4}
    \item \textbf{GPT-4o}~\cite{hurst2024gpt4o}
    \item \textbf{Chitrapathak} \footnote{\url{https://bit.ly/chitrapathak}}
    \item \textbf{Gemini-2.5 Pro}~\cite{google2025gemini2.5pro}
    \item \textbf{Gemini-2.5 Flash}~\cite{google2025gemini2.5flash}
\end{itemize}

The English model, trained exclusively on Latin-script documents does not transfer well to Indic scripts, achieving only 77.4\% accuracy on the ORB-Indic benchmark.

\subsection{Pipeline Integration}

To measure downstream impact, we insert our rotation classifier before OCR engines. We compare OCR related metrics with and without rotation correction. Field-level accuracy is used for structured forms like ORB-En (SROIE), and string alignment metrics are used for the free-text Indic dataset ORB-Indic, ORB-En (SynthDog) and ORB-En (FUNSD) data.

\begin{table}[htbp]
\resizebox{\linewidth}{!}{%

\begin{tabular}{lcc}
\toprule
\textbf{Model} 
& \textbf{ORB-En} 
& \textbf{ORB-Indic} \\
\midrule
TrOCR 
& 26.22 & \underline{68.90} \\

docTR 
& 63.28 & 46.95 \\
\hdashline

Nanonets-OCR-s 
& 24.83 & 37.50 \\

Gemma-3 27B 
& 30.57 & 33.23 \\

LLaMA-4 (Maverick) 17B 
& \underline{65.49} & 56.70 \\
\hdashline

GPT-4o 
& 59.58 & 50.60 \\

Gemini-2.5 Flash 
& 39.55 & 41.46 \\

Gemini-2.5 Pro 
& 34.11 & 40.85 \\
\hdashline

\textbf{Ours} 
& \textbf{96.81} & \textbf{92.68} \\
\bottomrule
\end{tabular}
}
\caption{Accuracy (\%) for 4-class document rotation classification on ORB-En and ORB-Indic. Best in bold, second-best underlined.}
\label{tab:4class-accuracy}
\end{table}

\section{Additional Ablation Study}
\label{appendix:ablation}

Additionally, we also conducted a second ablation study of the effect of the number of patches on the accuracy of rotation-class classification of the 12 class ORB-En FunSD dataset. This shows that while the accuracy increases as we increase the number of patches, it plateaus at around 32 patches, which has a latency of around 1.8 times the base dynamic cropping method, while yielding the same accuracy percentage, which makes the use of dynamic cropping a better option rather than specifying the patch size to be used explicitly. 

\begin{table}[htbp]
\centering
\resizebox{0.8\linewidth}{!}{%
\begin{tabular}{p{0.3\linewidth}c}
\toprule
\textbf{Patch count} & \textbf{Accuracy (\%) $\uparrow$} \\
\midrule
\makecell[l]{Single crop\\(1 patch)} & 60 \\
\makecell[l]{Moderate cropping\\(8 patches)} & 86 \\
\makecell[l]{Dense cropping\\(32 patches)} & 98 \\
\makecell[l]{\textbf{Dynamic cropping}\\\textbf{(Ours)}} & \textbf{98} \\
\bottomrule
\end{tabular}
}
\caption{Ablation study on the number of dynamic cropping patches for 12-class document rotation classification on the ORB-En FUNSD dataset.}
\label{tab:ablation-dynamic-crop-funsd}
\end{table}

\begin{table*}[htbp]
\centering
\resizebox{\textwidth}{!}{%
\begin{tabular}{lcccccc}
\toprule
\textbf{Model} 
& \multicolumn{2}{c}{\textbf{Ideal upright conditions}} 
& \multicolumn{2}{c}{\textbf{w/o Rotation}} 
& \multicolumn{2}{c}{\textbf{w/ Rotation}} \\
\cmidrule(lr){2-3} \cmidrule(lr){4-5} \cmidrule(lr){6-7}
& \textbf{Word $\downarrow$} & \textbf{Character $\downarrow$} 
& \textbf{Word $\downarrow$} & \textbf{Character $\downarrow$} 
& \textbf{Word $\downarrow$} & \textbf{Character $\downarrow$} \\
\midrule

\textit{ORB-En-SynthDog dataset — 4 class rotation} \\
docTR  & 56.23 & 40.04 & 88.03 & 69.07 & 56.23 & 40.04 \\
\hdashline
Nanonets-OCR-s        & \textbf{2.61} & \textbf{1.73} & \underline{44.14} & \underline{34.25} & \textbf{2.61} & \textbf{1.73} \\
Gemma-3 27B          & 61.56 & 30.29 & 75.15 & 50.06 & 61.56 & 30.29 \\
LLaMA-4 (Maverick) 17B & 29.37 & \underline{14.09} & 59.91 & 41.75 & 29.37 & \underline{14.09} \\
\hdashline
Qwen-2.5-VL           & 32.20 & 4.67 & 55.10 & 33.65 & 34.98 & 7.77 \\
mPLUG-DocOwl2         & 85.45 & 80.87 & 93.35 & 87.55 & 86.44 & 81.82 \\
\hdashline
GPT-4o               & 82.22 & 73.65 & 92.35 & 82.08 & 82.22 & 73.65 \\
Gemini-2.5 Pro       & 58.58 & 52.34 & 70.96 & 65.41 & 67.07 & 61.16 \\
Gemini-2.5 Flash     & \underline{22.43} & 15.33 & \textbf{26.24} & \textbf{18.66} & \underline{22.43} & 15.33 \\

\midrule

\textit{ORB-Indic dataset — 4 class rotation} \\
Nanonets-OCR-s       & 75.78 & 62.12 & 92.07 & 87.28 & 75.78 & 62.12 \\
Gemma-3 27B          & 75.43 & 48.19 & 92.01 & 75.53 & 76.16 & 49.41 \\
LLaMA-4 (Maverick) 17B & 47.25 & 24.73 & 84.41 & 68.83 & 47.25 & 24.73 \\
\hdashline
GPT-4o               & 82.43 & 63.76 & 94.55 & 89.14 & 82.00 & 64.14 \\
Chitrapathak         & \underline{34.56} & \underline{16.10} & 82.33 & 68.51 & \underline{34.56} & \underline{16.10} \\
Gemini-2.5 Pro       & 48.48 & 39.10 & \underline{67.50} & \underline{58.76} & 56.36 & 47.33 \\
Gemini-2.5 Flash     & \textbf{26.33} & \textbf{11.93} & \textbf{27.17} & \textbf{12.80} & \textbf{27.16} & \textbf{12.84} \\

\bottomrule
\end{tabular}}
\caption{ANLS (Average Normalized Levenshtein Similarity) under 4-class rotation across English and Indic datasets for different models. Word- and character-level scores are reported for ideal upright inputs, OCR without rotation correction, and OCR with rotation correction (lower is better).}
\label{tab:merged-ocr-anls-4class}
\end{table*}

\begin{table*}[htbp]
\centering
\resizebox{0.95\textwidth}{!}{%
\begin{tabular}{lcccccc}
\toprule
\textbf{Model} 
& \multicolumn{2}{c}{\textbf{Ideal upright conditions}} 
& \multicolumn{2}{c}{\textbf{w/o Rotation}} 
& \multicolumn{2}{c}{\textbf{w/ Rotation}} 
 \\
\cmidrule(lr){2-3} \cmidrule(lr){4-5} \cmidrule(lr){6-7}
& \textbf{WER $\downarrow$} & \textbf{CER $\downarrow$} 
& \textbf{WER $\downarrow$} & \textbf{CER $\downarrow$} 
& \textbf{WER $\downarrow$} & \textbf{CER $\downarrow$} 
\\
\midrule
\textit{ORB-En-SynthDog dataset} \\
docTR  & 0.583 & 0.411 & 0.896 & 0.694 & 0.583 & 0.411 \\
\hdashline
Nanonets-OCR-s      & \textbf{0.170} & 0.162 & 4.554 & 4.448 & \textbf{0.170} & 0.162 \\
Gemma-3 27B        & 0.605 & 0.317 & 0.754 & \underline{0.531} & 0.605 & 0.317 \\
LLaMA-4 (Maverick) 17B  & 0.296 & \underline{0.152} & 1.062 & 0.880 & 0.296 & \underline{0.152} \\
\hdashline
GPT-4o              & 0.840 & 0.786 & 0.936 & 0.852 & 0.840 & 0.786 \\
Gemini-2.5 Pro       & 0.636 & 0.595 & \underline{0.753} & 0.705 & 0.724 & 0.677 \\
Gemini-2.5 Flash      & \underline{0.230} & \textbf{0.146} & \textbf{0.267} & \textbf{0.179} & \underline{0.230} & \textbf{0.146} \\
\midrule
\textit{ORB-Indic dataset} \\
Nanonets-OCR-s      & 4.059 & 3.704 & 14.262 & 10.389 & 4.059 & 3.704 \\
Gemma-3 27B        & 0.751 & 0.523 & 0.959 & 0.832 & 0.756 & 0.532 \\
LLaMA-4 (Maverick) 17B  & 0.526 & 0.266 & 1.164 & 0.965 & 0.526 & 0.266 \\
\hdashline
GPT-4o              & 0.893 & 0.984 & 0.980 & 0.955 & 0.886 & 0.813 \\
Chitrapathak        & \underline{0.400} & \underline{0.181} & 2.913 & 1.746 & \underline{0.399} & \underline{0.180} \\
Gemini-2.5 Pro      & 0.501 & 0.425 & \underline{0.676} & \underline{0.617} & 0.566 & 0.500 \\
Gemini-2.5 Flash      & \textbf{0.251} & \textbf{0.113} & \textbf{0.262} & \textbf{0.117} & \textbf{0.272} & \textbf{0.129} \\
\bottomrule
\end{tabular}}
\caption{Performance comparison on ORB-En (SynthDog) and 4-class ORB-Indic documents with and without the proposed rotation correction pipeline, alongside ideal conditions. We report Word Error Rate (WER) and Character Error Rate (CER) on free-form outputs from both open-source and closed-source models (lower is better), computed using the HuggingFace \texttt{evaluate} function. Second best scores are underlined.}
\label{tab:merged-ocr-comparison-hf}
\end{table*}

\begin{table*}[htbp]
\centering
\resizebox{0.98\textwidth}{!}{%
\begin{tabular}{lcccccc}
\toprule
\textbf{Model} 
& \multicolumn{2}{c}{\textbf{Ideal upright conditions}} 
& \multicolumn{2}{c}{\textbf{w/o Rotation}} 
& \multicolumn{2}{c}{\textbf{w/ Rotation}} 
 \\
\cmidrule(lr){2-3} \cmidrule(lr){4-5} \cmidrule(lr){6-7}
& \textbf{Word $\uparrow$} & \textbf{Char $\uparrow$} 
& \textbf{Word $\uparrow$} & \textbf{Char $\uparrow$} 
& \textbf{Word $\uparrow$} & \textbf{Char $\uparrow$} 
\\
\midrule
\textit{ORB-En-SynthDog dataset} \\
docTR  & 0.496 & 0.441 & 0.197 & 0.160 &  0.496 & 0.441  \\
\hdashline
Nanonets-OCR-s        & \textbf{0.986} & \textbf{0.770} & \underline{0.612} & \underline{0.496} & \textbf{0.986} & \textbf{0.770}  \\
Gemma-3 27B        & 0.567 & 0.573 & 0.361 & 0.363 & 0.567 & 0.572  \\
LLaMA-4 (Maverick) 17B  & 0.816 & \underline{0.735} & 0.450 & 0.414 & 0.816 & \underline{0.735} \\
\hdashline
GPT-4o                & 0.252 & 0.216 & 0.162 & 0.153 & 0.252 & 0.216  \\
Gemini-2.5 Pro        & 0.462 & 0.391 & 0.328 & 0.284 & 0.363 & 0.314  \\
Gemini-2.5 Flash        & \underline{0.835} & 0.678 & \textbf{0.792} & \textbf{0.646} & \underline{0.835} & 0.678  \\
\midrule
\textit{ORB-Indic dataset} \\
Nanonets-OCR-s       & 0.266 & 0.230 & 0.211 & 0.188 & 0.266 & 0.230  \\
Gemma-3 27B        & 0.218 & 0.203 & 0.081 & 0.073 & 0.212 & 0.199  \\
LLaMA-4 (Maverick) 17B  & 0.489 & 0.460 & 0.164 & 0.147 & 0.489 & 0.460  \\
\hdashline
GPT-4o               & 0.119 & 0.106 & 0.118 & 0.105 & 0.134 & 0.119  \\
Chitrapathak         & \underline{0.661} & \underline{0.621} & 0.197 & 0.177 & \underline{0.661} & \underline{0.621}  \\
Gemini-2.5 Pro       & 0.527 & 0.522 & \underline{0.366} & \underline{0.330} & 0.476 & 0.440 \\
Gemini-2.5 Flash       & \textbf{0.810} & \textbf{0.770} & \textbf{0.790} & \textbf{0.750} & \textbf{0.810} & \textbf{0.770}\\
\bottomrule
\end{tabular}}
\caption{Performance comparison on ORB-En (SynthDog) and 4-class ORB-Indic documents with and without the proposed rotation correction pipeline, alongside ideal conditions. We report percentage match scores at word and character level (higher is better). Second best scores are underlined.}
\label{tab:merged-ocr-comparison}
\end{table*}

\end{document}